\documentclass[sigconf]{acmart}

\def\BibTeX{{\rm B\kern-.05em{\sc i\kern-.025em b}\kern-.08emT\kern-.1667em\lower.7ex\hbox{E}\kern-.125emX}}
    
\copyrightyear{2019} 
\acmYear{2019} 
\setcopyright{acmcopyright}
\acmConference[AIES '19]{AAAI/ACM Conference on AI, Ethics, and Society}{January 27--28, 2019}{Honolulu, HI, USA}
\acmBooktitle{AAAI/ACM Conference on AI, Ethics, and Society (AIES '19), January 27--28, 2019, Honolulu, HI, USA}
\acmPrice{15.00}
\acmDOI{10.1145/3306618.3314263}
\acmISBN{978-1-4503-6324-2/19/01}

\begin{document}

%
\title{Mapping Missing Population in Rural India: \\A Deep Learning Approach with Satellite Imagery}

%

\author{Wenjie Hu}
\affiliation{%
  \institution{Stanford University}}
\email{huwenjie@alumni.stanford.edu}

\author{Jay Harshadbhai Patel}
\affiliation{%
  \institution{Stanford University}}
\email{jayhp9@alumni.stanford.edu}

\author{Zoe-Alanah Robert}
\affiliation{%
  \institution{Stanford University}}
\email{zrobert7@alumni.stanford.edu}

\author{Paul Novosad}
\affiliation{%
  \institution{Dartmouth College}}
\email{paul.novosad@dartmouth.edu}

\author{Samuel Asher}
\affiliation{%
  \institution{World Bank}}
\email{sasher@worldbank.org}

\author{Zhongyi Tang}
\affiliation{%
  \institution{Stanford University}}
\email{zztang@stanford.edu}

\author{Marshall Burke}
\affiliation{%
  \institution{Stanford University}}
\email{mburke@stanford.edu}

\author{David Lobell}
\affiliation{%
  \institution{Stanford University}}
\email{dlobell@stanford.edu}

\author{Stefano Ermon}
\affiliation{%
  \institution{Stanford University}}
\email{ermon@cs.stanford.edu}

%
\renewcommand{\shortauthors}{Hu, et al.}

%
\begin{abstract}
Millions of people worldwide are absent from their country's census. Accurate, current, and granular population metrics are critical to improving government allocation of resources, to measuring disease control, to responding to natural disasters, and to studying any aspect of human life in these communities. Satellite imagery can provide sufficient information to build a population map without the cost and time of a government census. We present two Convolutional Neural Network (CNN) architectures which efficiently and effectively combine satellite imagery inputs from multiple sources to accurately predict the population density of a region. In this paper, we use satellite imagery from rural villages in India and population labels from the 2011 SECC census. Our best model achieves better performance than previous papers as well as LandScan, a community standard for global population distribution.
\end{abstract}

%
%


%

%
\maketitle

\section{Introduction}
In 2015, the United Nations set forth seventeen objectives to ``end poverty, protect the planet and ensure prosperity for all'' known as the Sustainable Development Goals (SGD) \cite{goals}. To monitor progress and ultimately achieve these objectives, accurate population statistics are essential. It is estimated that currently 300-350 million people worldwide are not included in their country's official population document, which hurts the measurement of SGD progress \cite{carr2013missing}. The ability to quickly and cost-effectively produce an accurate population map for a country has a multitude of benefits. Those missing populations are more likely to be marginalized communities which already do not receive sufficient resources from the government \cite{unicef}. An accurate population distribution is an essential basis for socioeconomic statistics, such as food, water, and energy demand in different regions of a country, which influence the policy-making and spending decisions of its government. Additionally, during natural disasters such as earthquakes and floods, an accurate population map can help organize rescue efforts more quickly and effectively. For regions with high infectious disease rates, a fine-grained population map also helps to prevent the spread of infectious diseases to locations with dense population \cite{Tatem2012,health}. 

However, creating a population map with high accuracy and high resolution is a challenging problem. Traditionally, it is done by performing a high-cost national census. The USAID Demographic and Health Survey (DHS) program performs surveys for developing countries typically every 5 years, and each survey costs anywhere from 1.1 million to 9.7 million USD \cite{doupe2016equitable}. The census surveys are even more expensive in developed countries like Europe, with a median cost of USD 5.57 per capita in 2010 \cite{uncensus}. For some countries with financial difficulties or political instability,  the census is carried out less frequently, as few as once every few decades \cite{decades,myanmar}. Reliance on out-of-date population statistics can lead to significant errors if used for policy making or resource allocation.

In this project, we aim to predict the population density of rural villages of India from high-resolution satellite imagery by utilizing Convolutional Neural Network (CNN) models. With the availability of high-frequency satellite images, we can predict population density every few days, saving the costs of on-site census surveys and avoiding the inaccuracies caused by the infrequency of census surveys. We demonstrate state-of-the-art prediction performance in villages of all states in India. By using satellite images with 10-30 meter resolution, our best models can predict aggregated village population in one Subdistrict (akin to a US county) with $R^2$ of 0.93, and individual village $\log_2$ population density with $R^2$ of 0.44. 

\section{Related Work}
\subsection{Traditional Methods}
Traditionally, population mapping is divided into two approaches, population projection and population disaggregation. Population projection predicts the future or current population of a region based on historical data. For most cases, simple linear regression is sufficient for the projection \cite{forecast}. In more complex models, projections take into consideration historic population data, birth rates, registered vehicles, etc \cite{Ratio-Correlation}. These models were used to project US county population in 5 years, which have very high accuracy with $R^2$ of 0.99. However, they don't provide information about the population distribution within each county.

\begin{figure}[ht]
\begin{center}
   \includegraphics[width=1.0\linewidth]{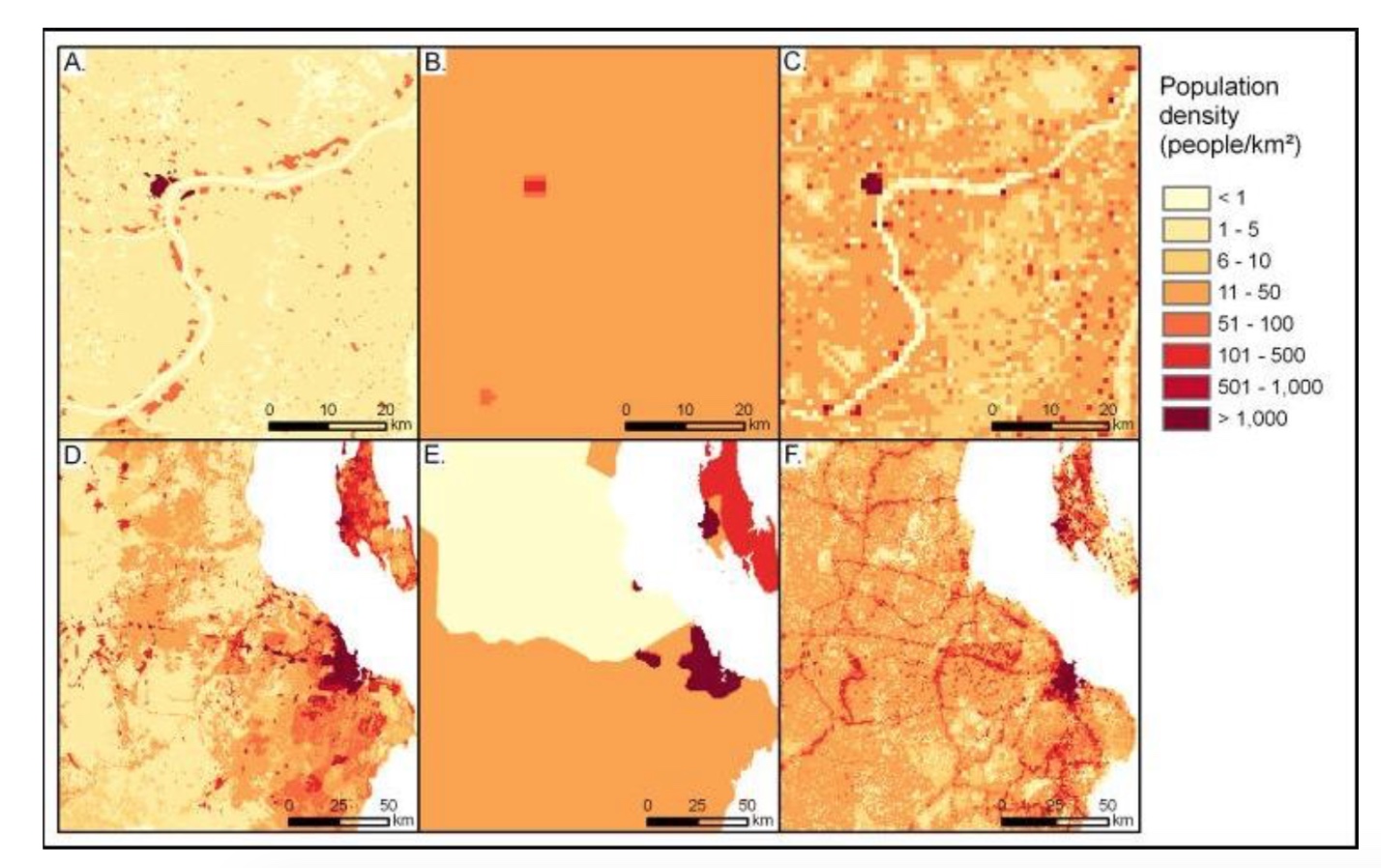}
\end{center}
   \caption{Population disaggregation visualization for WorldPop(A,D), GRUMP (B,E), LandScan (C,F). The upper 3 figures show a northeast region of Guinea along the Niger River; the lower 3 figures show the region around the largest city of Tanzania, Dar es Salaam. Figure from WorldPop (https://www.worldpop.org/)\cite{WinNT}.}
\label{figure1}
\end{figure}

The more challenging task is population disaggregation, which involves estimating the population distribution of a region given the total population. The most basic method is areal weighting/ interpolation, which assumes a uniform distribution across the region with a single population value \cite{areal}. The Gridded Population of the World (GPW) uses areal weighting with a resolution of 30 arc-seconds (approximately 1 km at the equator) \cite{CIESIN}. There are also many tools which implement a weighted surface for estimating a population's distribution, a technique called dasymetric weighting \cite{DBLP:journals/corr/abs-1708-09086}. The Global Rural Urban Mapping Project (GRUMP) uses nightlight imagery to add urban and rural boundaries to GPW~\cite{grump}. LandScan estimates the weighted surface (with 30 arc-seconds resolution) for population distribution based on land cover, roads, slope, urban areas, village locations~\cite{landscan}. AfriPop, AsiaPop, and AmeriPop are similar but for region-specific population disaggregation calculations, and they are combined in the 2013 WorldPop project \cite{worldpop}, which has a higher resolution of 100 meters. These disaggregation methods are compared visually in Figure \ref{figure1}.

\subsection{Machine Learning Methods}
In addition to the above traditional GIS approach, machine learning algorithms have been proposed in recent years to obtain better population disaggregation results. A random forest approach was used to estimate the population at 100m resolution for Vietnam, Cambodia, and Kenya, using features similar to LandScan \cite{stevens2015disaggregating}. The Facebook Connectivity Lab used a tailored CNN model to detect man-made structures from satellite imagery with 0.5m resolution, which achieved average precision of 0.95 and recall of 0.91. They then redistributed the population in GPW evenly to the areas covered by human-made structures, and create population maps with $\sim$30 meter resolution for 18 countries, not including India \cite{2017arXiv171205839T}.

Instead of disaggregation based on population estimates from census surveys, some CNN models are trained to estimate population directly from satellite imagery inputs. \citeauthor{doupe2016equitable} combined Landsat-7 satellite imagery with (DMSP/OLS) nighttime lights as CNN input, and predicted the log normalized population density for an area of 8km$^2$ (called \textbf{LL-raw}), where the ground-truth label was average population density of Sublocation (akin to a US county). The outputs were then converted into weights and used to create a weighted population density surface across the country with a single known total population (\textbf{LL-distributed}). The model was trained with 2002 Tanzanian Enumeration Areas and tested with 2009 Kenya Sublocations. During the test phase, the estimated population densities were averaged at the Sublocation level, and then compared to other methods. The results show that \textbf{LL-raw} has better accuracy than GRUMP and GWP estimates, and that \textbf{LL-distributed} outperforms a random forest model by 177\%~ on RMSE \cite{doupe2016equitable}. 

\citeauthor{DBLP:journals/corr/abs-1708-09086} in \citeyear{DBLP:journals/corr/abs-1708-09086} adopted a similar CNN approach to \citeauthor{doupe2016equitable}, but changed the model output from regression to classification of the power level of population. The model only uses Landsat imagery as input, and predicts population in the US with US Census Summary Grids data as ground-truth labels. The study divided the country into 15 regions, and trained an individual model for each region. The raw output feature vectors of the CNN model were first converted into population values for each input image, and then summed at the county level to produce \textbf{CONVRAW}. The outputs of the CNN were also fed into a second layer gradient boosting model to get an improved population estimate for each county called \textbf{CONVAUG}, where the census county population was used as labels. The results of CNN models achieved more than 0.9 $R^2$ against the ground truth, but still cannot perform better than the US government estimate based on historical census data \cite{DBLP:journals/corr/abs-1708-09086}.

\section{Data}
\subsection{Population Dataset}
\begin{figure}[ht]
\begin{center}
   \includegraphics[width=0.9\linewidth]{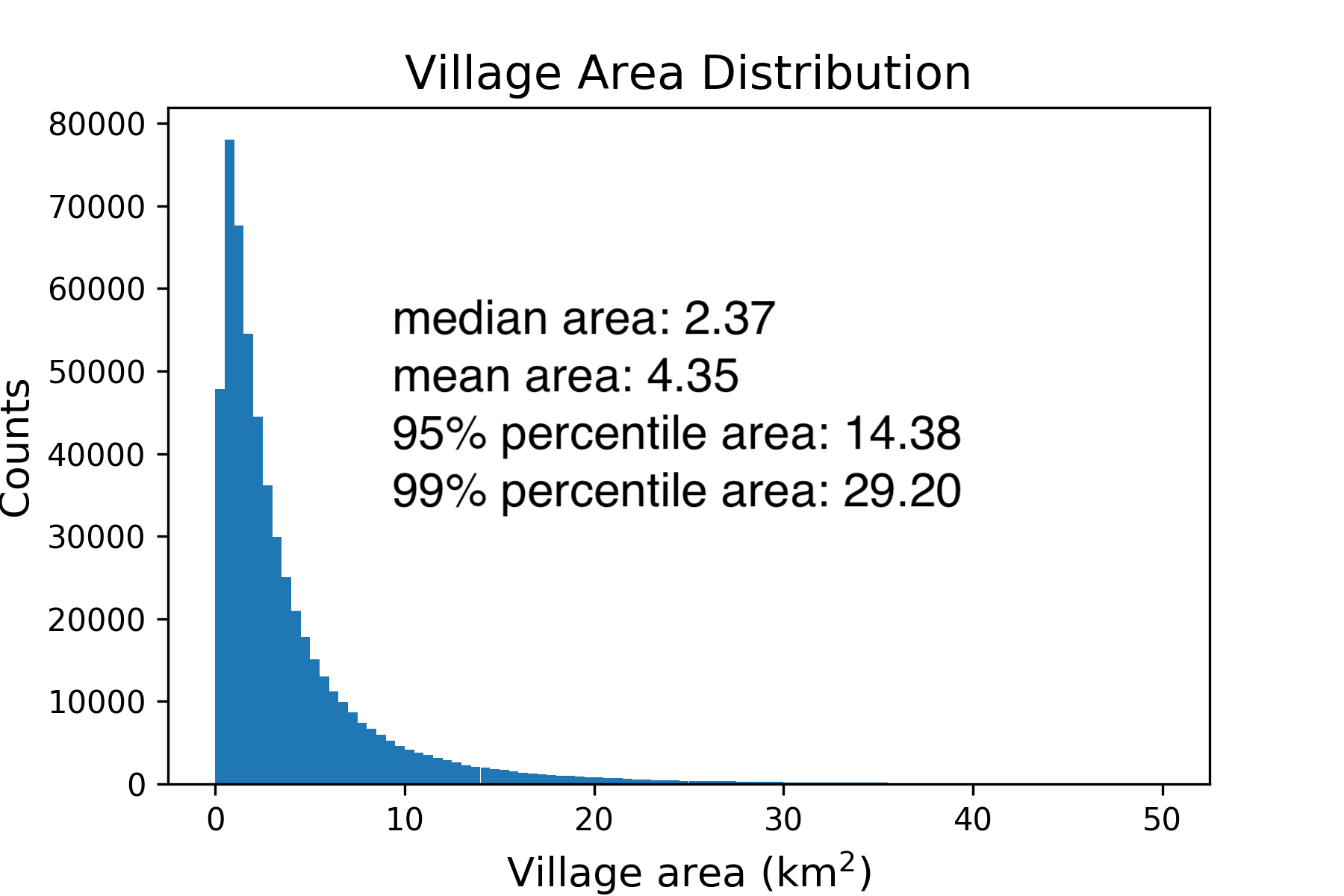}
\end{center}
   \caption{Village area distribution.}
\label{figure2}
\end{figure}

\begin{figure}[ht]
\begin{center}
   \includegraphics[width=0.9\linewidth]{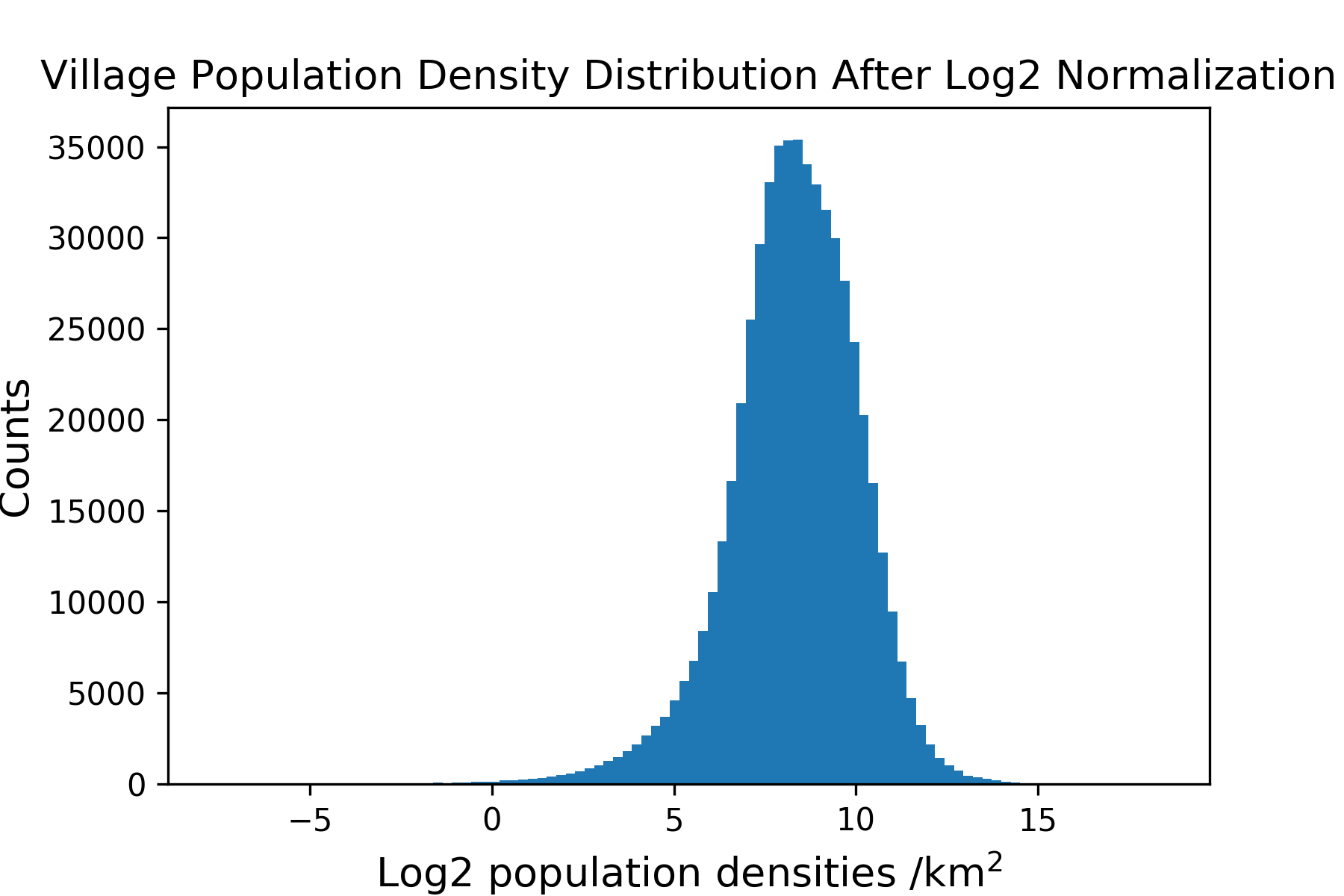}
\end{center}
   \caption{Village population density distribution after $\log_2$ normalization.}
\label{figure3}
\end{figure}

Our ground-truth Indian population dataset comes from a census survey Socio-Economic Caste Census in the year 2011 \cite{secc}. It includes more than 500,000 rural villages, covering 32 states, 619 districts, and 5724 sub-districts. It also provides the area of each surveyed village. Figure \ref{figure2} shows the distribution of areas follows a power law. These areas are used to calculate population density for each village. Similar to previous papers, we log normalize population density values with base 2, because most villages have small population density, and only a few have large density. The original density input may cause the model to have less ability to predict villages with higher population density. The distribution of villages density after $\log_2$ normalization is shown in Figure \ref{figure3}.

Inspecting the population datasets we observed some outlier values, such as a 100 km$^2$ area with just one person. We assume these are due to data collection and handling mistakes, therefore we remove 1\% of village data that had extreme population density values to prevent the model training from being affected by those outliers. More specifically, villages with top 0.5\% highest density and bottom 0.5\% lowest density were removed from the datasets. 

\subsection{Satellite Imagery}
\begin{figure}[ht]
\begin{center}
   \includegraphics[width=1.0\linewidth]{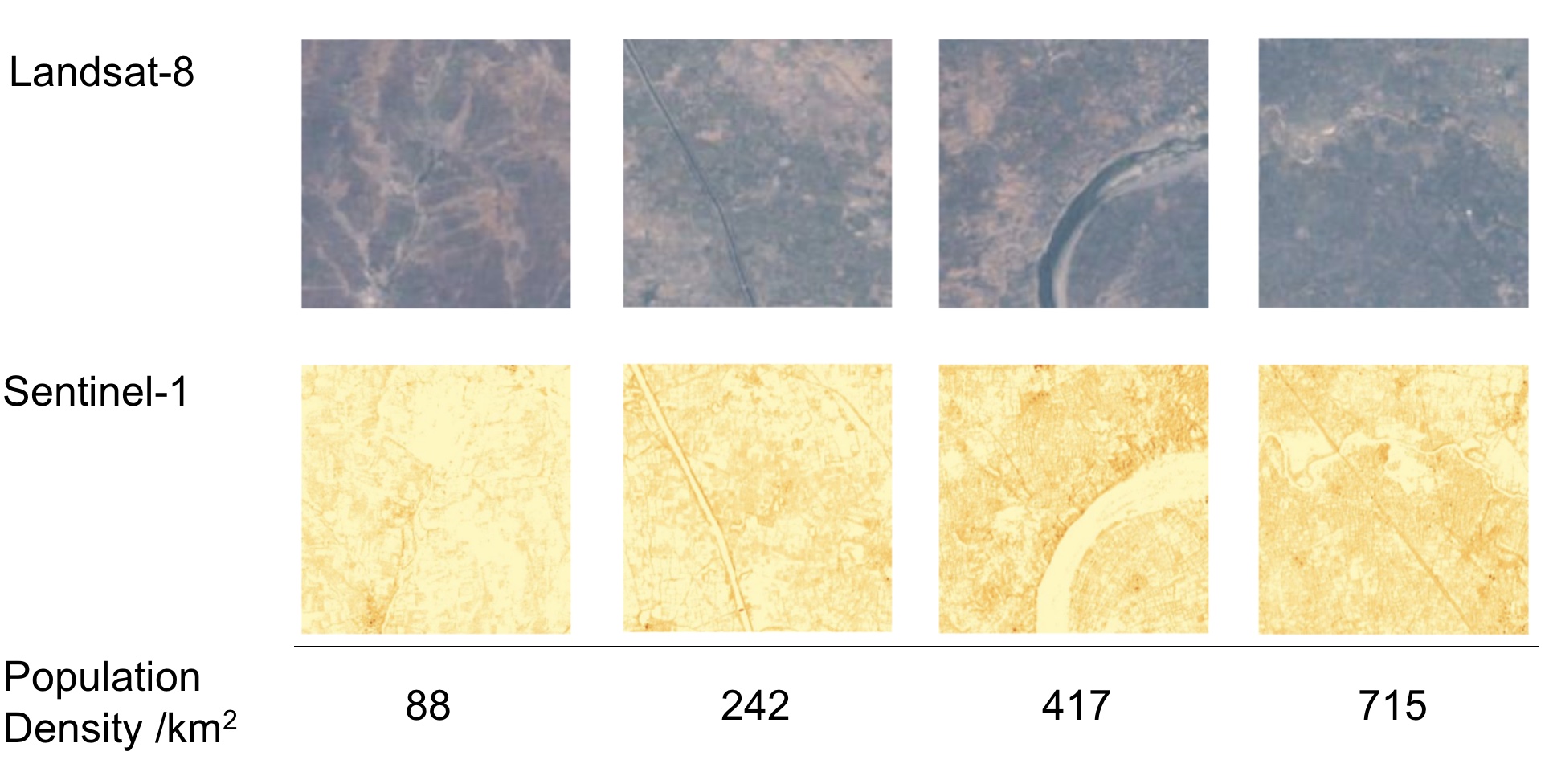}
\end{center}
   \caption{Landsat-8 and Sentinel-1 imagery examples, with corresponding ground truth population density from the survey.}
\label{figure4}
\end{figure}

For each village, we prepare one image from each satellite imagery source such that the village center is found at the image center. The village population depends on its area but our images have fixed size covering the same area, therefore we use population density as the output. 
We obtained 2 sets of satellite imagery from 2011, the same year the survey was conducted. The first set is from Landsat-8, an updated satellite from Landsat-7 whose images are also used by papers from \citeauthor{DBLP:journals/corr/abs-1708-09086} in \citeyear{DBLP:journals/corr/abs-1708-09086} and \citeauthor{doupe2016equitable} in \citeyear{doupe2016equitable}~\cite{landsat}. In contrast with previous papers which use most bands of Landsat, we use only Red, Green, and Blue (RGB) bands. The resulting images show the target regions in the same colors that humans see, and have 30-meter resolution. The second set is from Sentinel-1, a radar imaging satellite that measures ground surface reflectance, thereby capturing roads and roofs more accurately (due to their higher reflectance than natural land) \cite{sentinel}. This is a new dataset not used in the previously mentioned papers. Sentinel-1 images have 10-meter resolution, and the raw channel values are converted to visualized  RGB images to match Landsat-8. Both sets of images are converted to JPEG format from raw GeoTIFF files, which enables easy visualization and compression. Figure \ref{figure4} displays examples of these images. 

\begin{figure*}[ht]
\begin{center}
\includegraphics[width=0.8\linewidth]{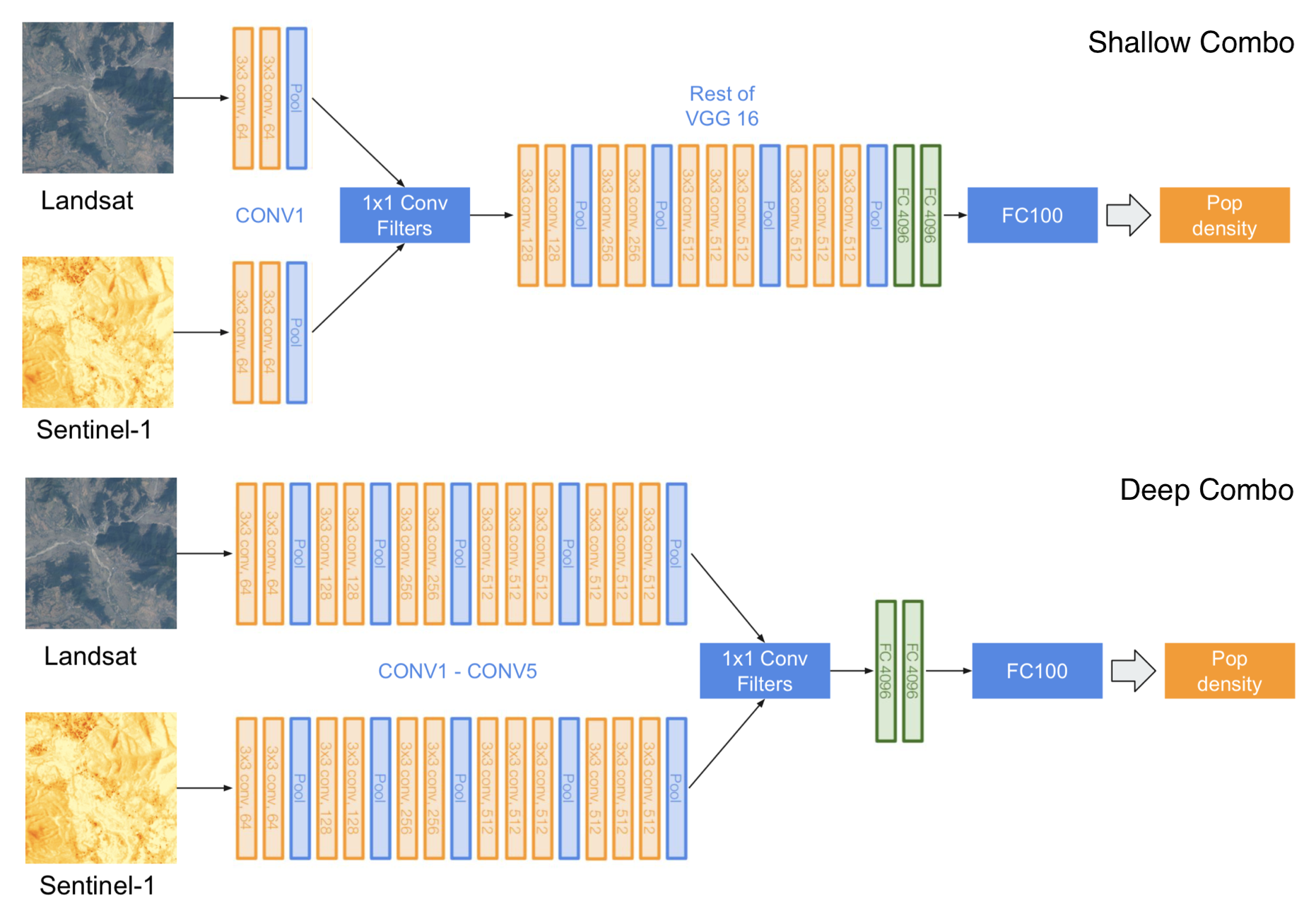}
\end{center}
   \caption{Custom CNN architecture \textbf{Shallow Combo} (upper) and \textbf{Deep Combo} (lower)}
\label{figure5}
\end{figure*}

Since Landsat-8 and Sentinel-1 have different spatial resolutions, they are cropped to cover the same area. Based on the village area distribution as shown in Figure \ref{figure2}, the covered area of images is determined to be 20.25 km$^2$ (4.5km $\times$ 4.5km), so that more than 95\% of villages are contained within a single satellite image. 

\subsection{Dataset Partition}
We split the dataset of ground-truth population (outputs) and images (inputs) pairs into 70\% training and 30\% validation partitions. We took additional measures to avoid the overlap of training images with validation images, which may affect the reliability of the validation split. Overlaps are very likely because the total area of all satellite images (over 10 million km$^2$) we obtained is much larger than the total area of India (about $3.3$ million km$^2$). To address this issue, we partition the data at the subdistrict level. We split all subdistricts into 4007 training and 1717 validation subdistricts. The training partition only has images of villages that belong to training subdistricts, and similarly for the validation partition. However, it is still possible that images overlap along the boundary of two adjacent subdistricts, contaminating the split. Thus, we remove additional images from the training partition. We say a pair of images overlap if the distance between their centers is closer than half of the height or width of the image (2.25km). Approximately an additional $5\%$ overlapping images are removed from the training partition. 

\section{Method}
We propose a deep learning approach that uses satellite images as inputs to a Convolutional Neural Network (CNN) to predict population density. The ground-truth population data is used as the label to train the model. We start with the VGG16 architecture~\cite{simonyan2014very}, using as input either Landsat-8 or Sentinel-1 RGB images. We use the implementation of VGG 16 from the TensorFlow Slim Library. To adapt the model for a regression problem, the model output size is set to be 1 for a single $\log_2$ population value. Additionally, the loss function in the model is changed to Mean Squared Error (MSE). Image inputs are resized to $224 \times 224 \times 3$ (width $\times$ height $\times$ channels), and we apply image augmentation including random cropping and flipping during training. The model weights are initialized with pre-trained weights from the ILSVRC-2010-CLS ImageNet classification dataset, omitting the last (classification) layer.

To fully utilize multiple satellite image sources in this study, we design two custom CNN architectures. The first custom architecture, called \textbf{Shallow Combo}, is shown in Figure 
\ref{figure5}. It is a modified version of VGG16 where Landsat-8 and Sentinel-1 images are input into the model separately as two branches. In each branch, two convolution layers and one max pooling layer (CONV1 in VGG16) are applied to each image. The two branches are then concatenated along the channel dimension. 1 $\times$ 1 convolutions filters are applied to halve the number of channels after concatenation, which fits the input shape of next convolutional layer. The rest of the VGG16 layers are applied to the merged branch. After the last fully connected layer of VGG16, one more fully connected layer of size 100 is added to the network. The layer is lastly used to predict the output $\log_2$ population density. During training, weights of all convolutional layers and the first two fully connected layers are initialized with pre-trained weights from the basic VGG16 architecture trained on ImageNet. 

The second custom CNN called \textbf{Deep Combo} is an improved version of Shallow Combo, also shown in Figure \ref{figure5}. The branches of Landsat-8 and Sentinel-1 are fed through all convolutional layers of VGG16, and concatenated along the channel dimension before the first fully connected layer. The same $1 \times 1$ convolutions filters are applied to obtain the same input shape before the fully connected layer. ImageNet weights are also used for initialization until the second fully connected layer.

\section{Experimental results}
\subsection{Evaluation Methods}
We evaluate each model on the validation set on two levels: raw village level and aggregated subdistrict level. The raw village level predictions are per-village $\log_2$ population density estimates from the model, which are compared with the ground-truth census population density of each village. This raw evaluation represents the most fine-grained comparison possible in our dataset. At the aggregated subdistrict level, the raw per-village density predictions are converted to the total village population using the village area from the survey. The village populations within each subdistrict are aggregated to a total population for that subdistrict. Evaluations at the aggregate levels are performed in previous papers, which we are going to compare with. In short, the village level evaluation uses $\log_2$ population density values, while subdistrict level evaluation uses real population numbers.  

In terms of evaluation metrics, we compare different models mainly based on $R^2$ and Pearson Correlation. $R^2$ measures how much true variance is captured by the model, and attains a perfect value at 1. If a model makes constant average predictions, it will have $R^2$ score of 0. Pearson Correlation measures the linear correlation between the predicted and true values, implying a total positive correlation when +1 and a total negative correlation when -1. We also use other metrics to compare results with previous papers, such as MAPE (Mean Absolute Percentage Error), \%RMSE (percent Root Mean Squared Error). 

\subsection{Baseline: LandScan}
To establish a baseline comparison for this project, we compared our ground truth population densities with those of LandScan, which is considered a community standard for global population distribution \cite{landscan}. LandScan is a grid containing a numerical population estimate for every cell of size 30x30 arc-second in earth coordinates, which is around 1km on the equator (smaller at different latitudes). This estimate represents an average (over 24 hours), or ambient population distribution (not just sleeping location). For a single village, we take a square of 30 arc-second cells which most closely matches the villages area, as well as the village latitude and longitude. The population density of the village is calculated by dividing the LandScan population estimates within the cells with the area covered by the cells.  

We use this method to obtain the LandScan estimate of population density for all villages in our validation dataset. The evaluation results are shown as \textbf{LandScan} model in Table \ref{table2}. It has good performance in estimating aggregated population on subdistrict level, but it doesn't perform well on per-village level. It is reasonable because LandScan is a traditional population disaggregation method using only the general features on the maps, such as rivers, roads, land cover types. It does not have fine-grained information as our models have from satellite imagery. 

\subsection{Single State Training}
Due to the large amount of data, we initially generate a smaller dataset with villages in a single state, Gujarat. The small dataset contains around 13,000 villages, which allows us to compare different models quickly, and to tune the hyper-parameters of the CNN models. We experiment with different input sources and the different CNN architectures described above.  We train the basic VGG-16 separately with Landsat-8 (\textbf{L8}) and Sentinel-1(\textbf{S1}) inputs. The L8 and S1 inputs are also concatenated along the image channels and feed into VGG-16 as single input (\textbf{S1L8-Concat}). Furthermore, \textbf{Shallow Combo} and \textbf{Deep Combo} architectures are used to process two type of satellite images separately. We use NVIDIA Tesla P100 GPU with 16G memory from Google Compute Engine for training. It sets the limit of input batch size to 48 for the Deep Combo architecture, thus this batch size is also used for other models to ensure a fair comparison. With several iterations, the optimal hyper-parameters are found as follows: learning rate: $10^{-5}$, exponential learning decay rate: $10^{-1}$, weight decay: 5x10$^{-3}$, dropout: 0.8.

\begin{table}[ht]
  \caption{Evaluation of Single State Training}
  \centering
  \begin{tabular}{l | c c | c c}
    \toprule  
    Model & \multicolumn{2}{c|}{Village level} & \multicolumn{2}{c}{Subdistrict level} \\
    & $R^2$ & Pearson & $R^2$ & Pearson \\
    \hline\hline
    L8 			& 0.111 & 0.443 & 0.597 & 0.775 \\
    S1 			& 0.120 & 0.425 & 0.701 & 0.839 \\
    L8S1-Concat & 0.111 & 0.472 & 0.305 & 0.764 \\
    Shallow Combo& \textbf{0.200} & 0.488 & 0.739 & 0.889 \\
    Deep Combo   & 0.167 & \textbf{0.489} & \textbf{0.772} & \textbf{0.921} \\
    \bottomrule
  \end{tabular}
  \label{table1}
\end{table}

The evaluations of the above 5 models are shown in Table \ref{table1}. In general, the models have lower accuracies when predicting per-village level population densities, but the accuracy significantly improves when the individual predictions are aggregated for the subdistrict level evaluation. The \textbf{S1} model has better performance than the \textbf{L8} model, possibly due to its high resolution as well as that measuring ground reflectance by radar is more effective. However, simply concatenating L8 and S1 inputs in \textbf{L8S1-Concat} does not lead to further improvements, resulting in performance slightly worse than using S1 alone. This is likely due to training difficulties when using a combination of L8 and S1 images. In contrast, our \textbf{Shallow Combo} and \textbf{Deep Combo} architectures have significantly better performance.
These results indicate that when using inputs from different sources, the CNN model needs to process them separately to extract their semantic features, so that useful information from both sides can be utilized.

\subsection{All States Training}
After training on a single state, we move on to train on around 350,000 villages in all 32 states in India. \textbf{L8}, \textbf{S1}, \textbf{Shallow Combo} and \textbf{Deep Combo} are trained respectively with the same hyper-parameters found when training in the single state case. The evaluation results are summarized in Table \ref{table2}.  With more data provided, all the models show significant improvements. All models exceed the performance of the  \textbf{LandScan} baseline. \textbf{Shallow Combo} and \textbf{Deep Combo} models still outperform the basic VGG-16 models with a single image source. By comparing between two custom architectures, \textbf{Shallow Combo} has better performance on predicting population on a single village, while \textbf{Deep Combo} captures more general information in the region and has better accuracy when predicting aggregated population at the subdistrict level.

\begin{table}[ht]
  \caption{Evaluation of All States Training}
  \centering
  \begin{tabular}{l | c c | c c}
    \toprule  
    Model & \multicolumn{2}{c|}{Village level} & \multicolumn{2}{c}{Subdistrict level} \\ 
    & $R^2$ & Pearson & $R^2$ & Pearson \\
    \hline\hline
    L8 			& 0.346 & 0.596 & 0.838 & 0.919 \\
    S1 			& 0.327 & 0.597 & 0.890 & 0.944 \\
    Shallow Combo& \textbf{0.438} & \textbf{0.663} & 0.906 & 0.954 \\
    Deep Combo	& 0.389 & 0.645 & \textbf{0.931} & \textbf{0.965} \\
    \hline
    LandScan    & -0.553& 0.476 & 0.835 & 0.928 \\
    CONVRAW    & 0.322 & 0.592 & 0.850 & 0.937 \\
    \bottomrule
  \end{tabular}
  \label{table2}
\end{table}

\subsection{Comparison with Prior Works}
In this section, we compare our models with others from prior studies, using aggregated evaluation metrics. Table \ref{table3} summarizes the reported accuracies from the papers from \citeauthor{doupe2016equitable} in \citeyear{doupe2016equitable} with study area in Kenya, and from \citeauthor{DBLP:journals/corr/abs-1708-09086} in 2017 with study area in the United States. Since we do not use subdistrict level census population for either training or prediction improvement, to ensure the fair comparison, we compare to the models in the papers also without the assistance of additional data from the aggregated level, namely \textbf{LL-raw} from Doupe's paper and \textbf{CONVRAW} from Robinson's paper. Comparing with Robinson's paper shows that even though we use a smaller (average) aggregation area in India, we still achieve significantly better $R^2$ and MAPE. The aggregation area in Doupe's study is even smaller, but their model also has much higher errors compared to ours.

\begin{table}[ht]
  \caption{Comparison with Prior Works}
  \centering
  \begin{tabular}{l | p{1.7cm} p{1.9cm} p{1.7cm}}
    \toprule  
    Paper & Ours & Robinson 2017 & Doupe 2016 \\
    \hline\hline
    Study area 	& India  & US & Kenya \\
    Aggr. area & 424 km$^2$ & 2584 km$^2$ & 88 km$^2$ \\
    Model & Deep Combo & CONVRAW & LL-Raw  \\
    $R^2$ 	& 0.931 & 0.910 & -  \\
    MAPE    & 21.5& 73.8 & -  \\
    \%RMSE    & 24.3 & - & 145.43  \\
    \bottomrule
  \end{tabular}
  \label{table3}
\end{table}

However, it is difficult to compare models trained and tested across different countries. To address this problem, we use the same methods from \citeauthor{DBLP:journals/corr/abs-1708-09086}, and apply them to our India study area. Specifically, we trained a classification model using a single Landsat input. The categorical outputs of the CNN are in the bins of $\log_2$ population density values for each Indian village. Following its \textbf{CONVRAW} approach, we convert the predicted class to population density, and aggregate the population at the subdistrict level. The predictions are still not as good as most of our models, which are shown as \textbf{CONVRAW} in Table \ref{table2}. Its performance is close to our L8 model, which differs from it by using regression instead of classification. The comparison shows the regression model is better at predicting fine-grained population while classification is better for aggregate predictions.

\begin{figure*}[!ht]
\begin{center}
\includegraphics[width=1.0\linewidth]{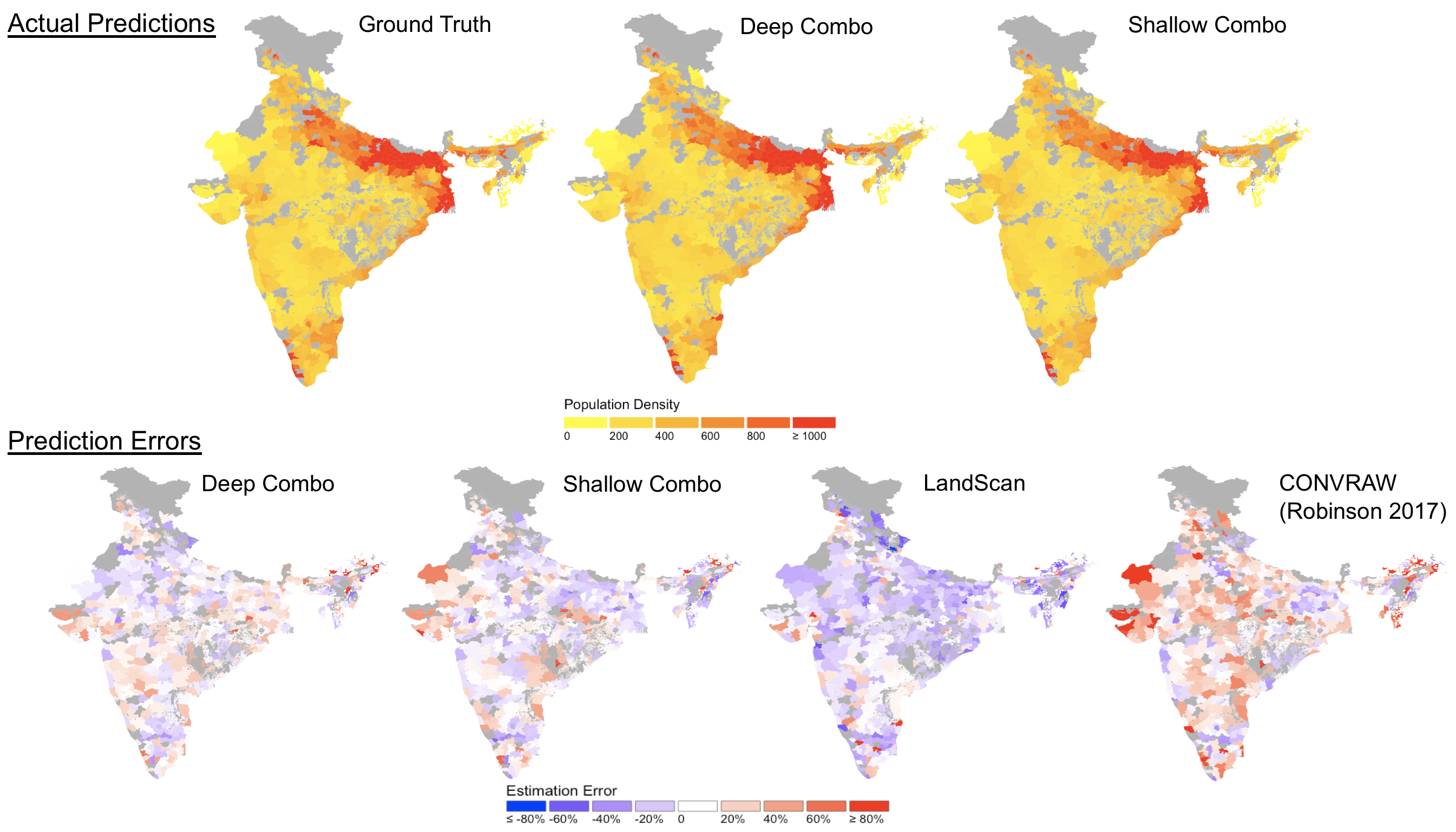}
\end{center}
   \caption{Population mapping visualization on the validation dataset. The upper row shows the population predictions of Deep Combo and Shallow Combo model compared to the ground-truth values. The lower row shows the prediction errors of Deep Combo and Shallow Combo compared with the baseline LandScan, and CONVRAW model from a previous paper \cite{DBLP:journals/corr/abs-1708-09086}. The grey areas are the regions with no validation data.}
\label{figure6}
\end{figure*}

Finally, we visualize the population mapping of different models in districts of India, the higher administrative level than subdistrict,  as shown in Figure \ref{figure6}. It includes 533 districts that contain evaluation results from 1717 validation subdistricts. The visualization includes our \textbf{Deep Combo} and \textbf{Shallow Combo} model predictions compared with ground-truth population density distribution across districts in India. It also compares the prediction errors of the two models with those of the baseline \textbf{LandScan} and \textbf{CONVRAW} model from \citeauthor{DBLP:journals/corr/abs-1708-09086} Overall, both \textbf{Deep Combo} and \textbf{Shallow Combo} models map the rural village population of India very close to the true distribution. The prediction errors map further reveals that Deep Combo has better performance on high level aggregation than other models, while \textbf{LandScan} tends to underestimate the population and \textbf{CONVRAW} tends to overestimate. 

\section{Conclusion}
The results of our CNN models on the population density prediction directly from satellite imagery are very promising. We see a higher accuracy than previous methods on both per-image raw level and aggregated region level. We attribute the success to the following reasons:
\begin{itemize}
\item Additional Sentinel-1 satellite imagery that provides more information on human settlements. 
\item Custom \textbf{Shallow Combo} and \textbf{Deep Combo} CNN architectures that better utilize different imagery sources.
\item A larger and fine-grained dataset for CNN training.
\item Better computational resources that enable larger image input size and batch size.
\end{itemize}
Our study largely achieves the initial goal of producing accurate population estimates enabling population mapping directly from satellite imagery, especially for rural areas. The results show that population prediction at a relatively coarse scale (such as district or subdistrict level) is quite accurate, while prediction at the village level directly from satellite imagery remains challenging. However, our village level predictions already have significantly better performance than traditional methods such as LandScan, and may still have room for improvement if images with higher resolution are available. Our population mapping models are likely useful for assisting governments in better providing for their citizens, improving resource allocation in natural disasters, aiding in infectious disease tracking, and reducing bias in progress measurement for the United Nations SDGs.

%
\begin{acks}
This work was supported by Data for Development Initiative at the Stanford Center on Global Poverty and Development. All data in this work was processed and shared by Paul Novosad (Dartmouth College) and Sam Asher (World Bank). 
\end{acks}

%
\bibliographystyle{ACM-Reference-Format}
\bibliography{reference}

%


\end{document}